\documentclass[final]{cvpr}

\usepackage{times}
\usepackage{epsfig}
\usepackage{graphicx}
\usepackage{amsmath}
\usepackage{amssymb}
\usepackage{bm}
\usepackage{mathrsfs}
\usepackage{algorithm}
\usepackage{algorithmic}
\usepackage{multirow}
\usepackage{eqparbox}
\usepackage{color}
\usepackage{makecell}

\usepackage[pagebackref=true,breaklinks=true,colorlinks,bookmarks=false]{hyperref}

\begin{document}
	
	\title{HCRF-Flow: Scene Flow from Point  Clouds with Continuous High-order CRFs and Position-aware Flow Embedding}
	
\author{Ruibo Li$^{1,2}$, ~~ Guosheng Lin$^{1,2}$\thanks{Corresponding author: G. Lin. (e-mail:{ $\tt gslin@ntu.edu.sg$ })}, ~~ Tong He$^3$, ~~ Fayao Liu$^4$, ~~Chunhua Shen$^3$\\
	$^{1}$S-Lab, Nanyang Technological University, Singapore \\
	$^{2}$School of Computer Science and Engineering, Nanyang Technological University, Singapore \\
	$^{3}$University of Adelaide, Australia \
	$^{4}$Institute for Infocomm Research, A*STAR, Singapore\\ 
	E-mail: { $\tt ruibo001@e.ntu.edu.sg$ }, { $\tt gslin@ntu.edu.sg$ }
}
	
	\maketitle

	\begin{abstract}
		Scene flow in 3D point clouds plays an important role in understanding dynamic environments. Although significant advances have been made by deep neural networks, the performance is far from satisfactory as only per-point translational motion is considered, neglecting the constraints of the rigid motion in local regions. To address the issue, we propose to introduce the motion consistency to force the smoothness among neighboring points. In addition, constraints on the rigidity of the local transformation are also added by sharing unique rigid motion parameters for all points within each local region. To this end, a high-order CRFs based relation module (Con-HCRFs) is deployed to explore both point-wise smoothness and region-wise rigidity. To empower the CRFs to have a discriminative unary term, we also introduce a position-aware flow estimation module to be incorporated into the Con-HCRFs. Comprehensive experiments on FlyingThings3D and KITTI show that our proposed framework (HCRF-Flow) achieves state-of-the-art performance and significantly outperforms previous approaches substantially. 
	\end{abstract}
	
	\section{Introduction}		
	
	Scene flow estimation~\cite{vedula1999three} aims to provide dense or semi-dense 3D vectors, representing the per-point 3D motion in two consecutive frames. The information provided has proven invaluable in analyzing dynamic scenes. 
	Although significant advances have been made in the 2D optical flow, the counterpart in 3D point cloud is far more challenging. This is partly due to the irregularity and sparsity of the data, but also due to the diversity of the scene.

		\begin{figure}[ht]
		\centering
		\includegraphics[height=7.5cm]{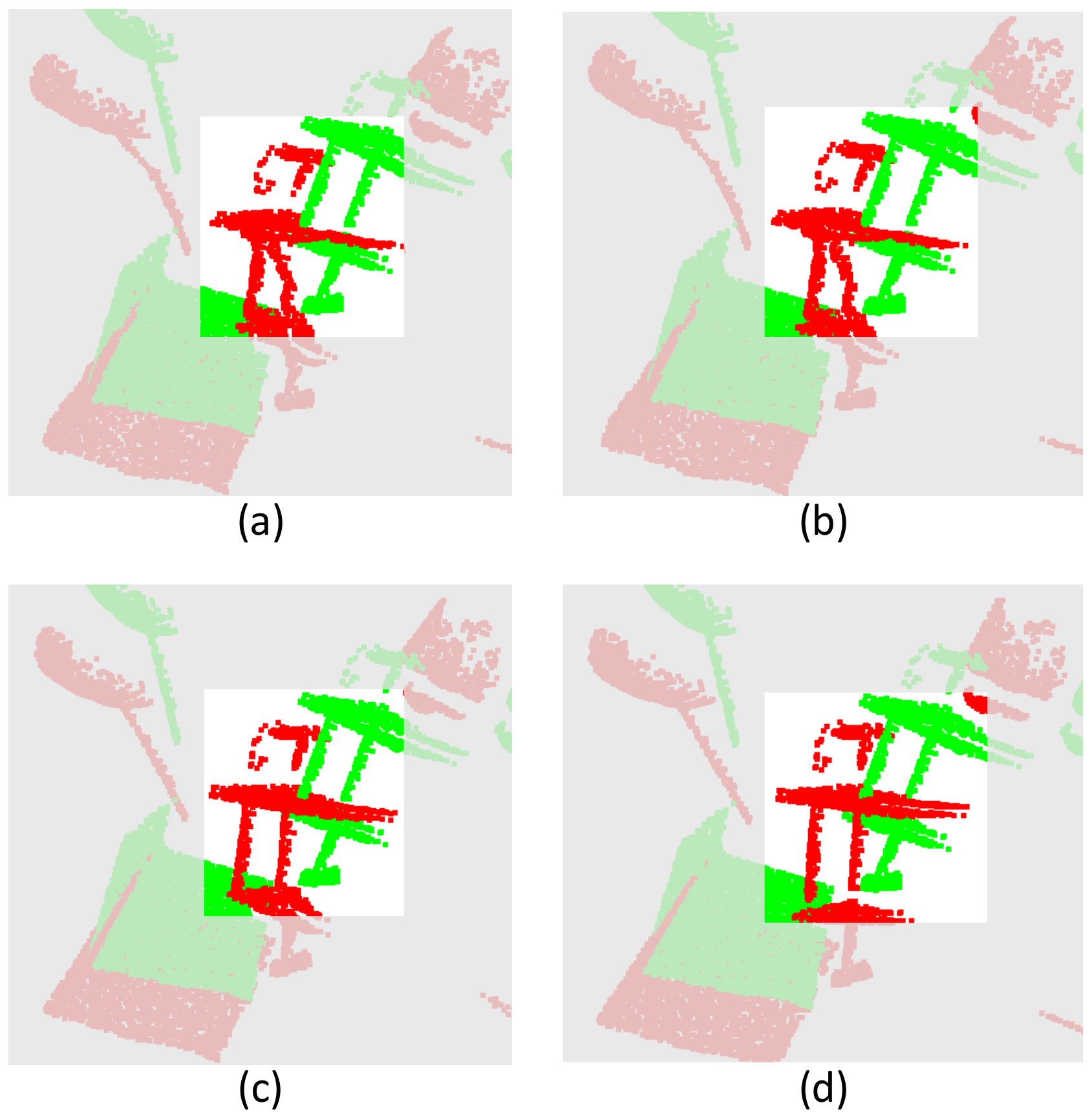}
		\caption{The warped point cloud at the next frame based on different scene flow. Green points represent the point cloud at frame~$t$. Red points are the warped results at frame $t+1$ by adding scene flow back to corresponding green points.
			(a) scene flow produced by FlowNet3D~\cite{liu2019flownet3d}; (b) scene flow produced by FlowNet3D and refined by a conventional CRF; (c) scene flow produced by FlowNet3D and refined by our continuous high order CRFs; (d) ground truth scene flow.
			The local structure of the warped point cloud is distorted in Flownet3d and the conventional CRF but preserved in our method.
		}
		\label{fig1}	
	\end{figure}

	As pointed out in~\cite{man1982computational}, most of the structures in the visual world are rigid or at least nearly so.
	Many previous top-performing approaches \cite{liu2019flownet3d, gu2019hplflownet, wu2020pointpwc, puy2020flot} simplify this task as a regression problem by estimating a point-wise translational motion. Although promising results have been achieved, the performance is far from satisfactory as the potential rigid motion constraints existing in the local region are ignored. As shown in Fig.~\ref{fig1}(a), the results generated by the FlowNet3D~\cite{liu2019flownet3d} are deformed and fail to maintain the local geometric smoothness. A straightforward remedy is to utilize pair-wise regularization to smooth the flow prediction. However, ignoring the potential rigid transformations makes it hard to maintain the underlying spatial structure, as presented in Fig.~\ref{fig1}(b).

	To address this issue, we propose a novel framework termed HCRF-Flow, which consists of two components: a position-aware flow estimation module (PAFE) for per-point translational motion regression and a continuous high-order CRFs module (Con-HCRFs) for the refinement of the per-point predictions by considering both spatial smoothness and rigid transformation constraints. 
	Specifically, in Con-HCRFs, a pairwise term is designed to encourage neighboring points with similar local structure to have similar motions. In addition, a novel high order term is designed to encourage each point in a local region to take a motion obeying the shared rigid motion parameters, i.e., translation and rotation parameters, in this region.

	In point cloud scene flow estimation, it is challenging to aggregate the matching costs, which are calculated by comparing one point with its softly corresponding points.
	To encode this knowledge into the embedding features, we propose a position-aware flow embedding layer in the PAFE module. 
	In the aggregation step, we introduce a pseudo matching pair that is applied to calculate the difference of the matching cost.
	For each softly corresponding pair, both its position information and the matching cost difference will be considered to output weights for aggregation.

	Our main contributions can be summarized as follows:
	\begin{itemize}
		\item  
		We propose a novel scene flow estimation framework HCRF-Flow by combining the strengths of DNNs and CRFs to perform a per-point translational motion regression and a refinement with both pairwise and region-level regularization;
		\item  
		Formulating the rigid motion constraints as a high order term, we propose continuous high-order CRFs (Con-HCRFs) to model the interaction of points by imposing point-level and region-level consistency constraints.

		\item  
		We present a novel position-aware flow estimation layer to build reliable matching costs and aggregate them based on both position information and matching cost differences. 
		
		\item  
		Our proposed HCRF-Flow significantly outperforms the state-of-the-art on both FlyingThing3D and KITTI Scene Flow 2015 datasets.
		In particular, we achieve Acc3DR scores of 95.07\% and 94.44\% on FlyingThing3D and KITTI, respectively.
	\end{itemize}
	
	\subsection{Related work}
	
	\noindent\textbf{Scene flow from RGB or RGB-D images}\quad 
	Scene flow is first proposed in~\cite{vedula1999three} to represent the three-dimensional motion field of points in a scene.
	Many works~\cite{huguet2007variational,pons2007multi,sun2015layered,valgaerts2010joint,vogel20113d,vogel2013piecewise,vogel20153d,menze2015object,ma2019deep,quiroga2014dense,hornacek2014sphereflow} try to recover scene flow from stereo RGB images or monocular RGB-D images.
	The local rigidity assumption has been applied in scene flow estimation from images.
	\cite{vogel2013piecewise,vogel20153d,menze2015object,ma2019deep} directly predict the rigidity parameter of each local region to produce scene flow estimates.
	\cite{vogel20113d,quiroga2014dense,hornacek2014sphereflow} add a rigidity term into the energy function to constrain the scene flow estimation. 
	Compared with them, our method is different in the following aspects: 
	1) our method formulates the rigidity constraint as a high order term in Con-HCRFs. It encourages the region-level rigidity of point-wise scene flow rather than directly computing rigidity parameters.
	Thus, our Con-HCRFs can be easily added to other point cloud scene flow estimation methods as a plug-in module to improve the rigidity of their predictions;
	2) our method targets irregular and unordered point cloud data instead of well organized 2D images.

	\noindent\textbf{Deep scene flow from point clouds}\quad 
	Some approaches~\cite{dewan2016rigid,ushani2017learning} estimate scene from point cloud via traditional techniques. 	
	Recently, inspired by the success of deep learning for point clouds, more works~\cite{gu2019hplflownet,behl2019pointflownet,liu2019flownet3d,wu2020pointpwc,puy2020flot,mittal2020just,liu2019meteornet,wang2018deep} have employed DNNs in this field.
	\cite{liu2019flownet3d} estimates scene flow based on PointNet++~\cite{qi2017pointnet++}.
	\cite{gu2019hplflownet} proposes a sparse convolution architecture for scene flow learning, \cite{wu2020pointpwc} designs a coarse-to-fine scene flow estimation framework, and \cite{puy2020flot} estimates the point translation by point matching. 
	Despite achieving impressive performance, these methods neglect the rigidity constraints and estimate each point's motion independently.
	Although the rigid motion for each point is computed in~\cite{behl2019pointflownet},  
	the  per-point rigid parameter is independently regressed by a DNN without fully considering its geometry constraints.
	Unlike previous methods, we design a novel Con-HCRFs to explicitly model both spatial smoothness and rigid motion constraints.
	
	\noindent\textbf{Deep learning on 3D point clouds}\quad 
	Many works~\cite{qi2017pointnet,qi2017pointnet++,thomas2019kpconv,mao2019interpolated,wu2019pointconv,liu2019relation,wang2019graph,hu2020randla} focus on learning directly on raw point clouds.
	PointNet~\cite{qi2017pointnet} and  PointNet++~\cite{qi2017pointnet++} are the pioneering works which use shared Multi-Layer-Perception (MLP) to extract features and a max pooling to aggregate them.
	\cite{hu2020randla,wang2019graph} use the attention mechanism to produce aggregation weights.
	\cite{hu2020randla,liu2019relation} encode shape features from local geometry clues to improve the feature extraction.
	Inspired by these works, we propose a position-aware flow embedding layer to dynamically aggregate matching costs based on  both position representations and matching cost differences for better matching cost aggregation.

	\noindent\textbf{Conditional random fields (CRFs)}\quad 
	CRFs are a type of probabilistic graphical models, which are widely used to model the effects of interactions among examples in numerous vision tasks~\cite{krahenbuhl2011efficient,liu2015deep,chen2017deeplab,zhang2019canet,xu2017multi}.
	In point cloud processing, previous works~\cite{yang2019learning,tchapmi2017segcloud,choy20194d} apply CRFs to discrete labeling tasks for spatial smoothness.
	In contrast to the CRFs in these works, the variables of Con-HCRFs are defined in a continuous domain, and two different relations are modeled by Con-HCRFs at the point level and the region level.

	\begin{figure*}[tb]
		\centering
		\includegraphics[height=4.5cm]{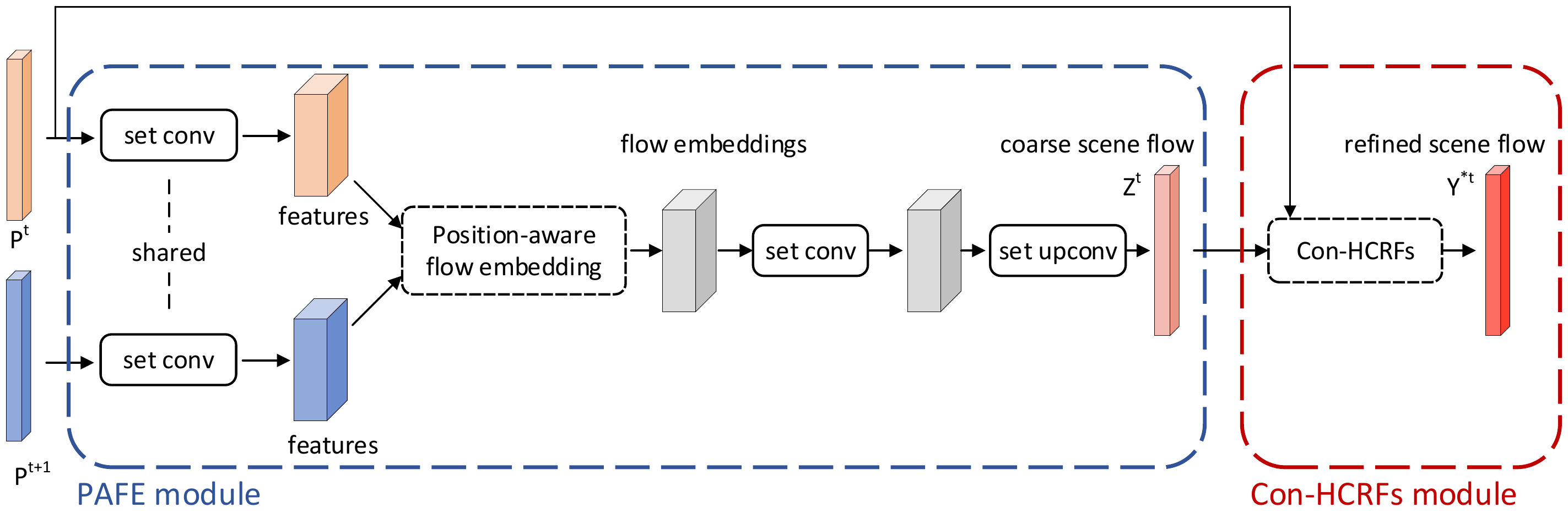}
		\caption{HCRF-Flow architecture. 
			Our HCRF-Flow consists of two components: a PAFE module to produce per-point initial scene flow and a Con-HCRFs module to refine the initial scene flow.
			Our proposed position-aware flow embedding layer is employed in the PAEF module to encode motion information.  
			We build two different architectures of the PAFE module: one is designed by considering only single-scale feature (similar to FlowNet3D~\cite{qi2017pointnet}). The other one introduces a pyramid architecture (similar to PointPWC-Net~\cite{wu2020pointpwc}).
		}
		\label{fig_a}	
	\end{figure*}
	
	\begin{figure}[tb]
		\centering
		\includegraphics[height=2.5cm]{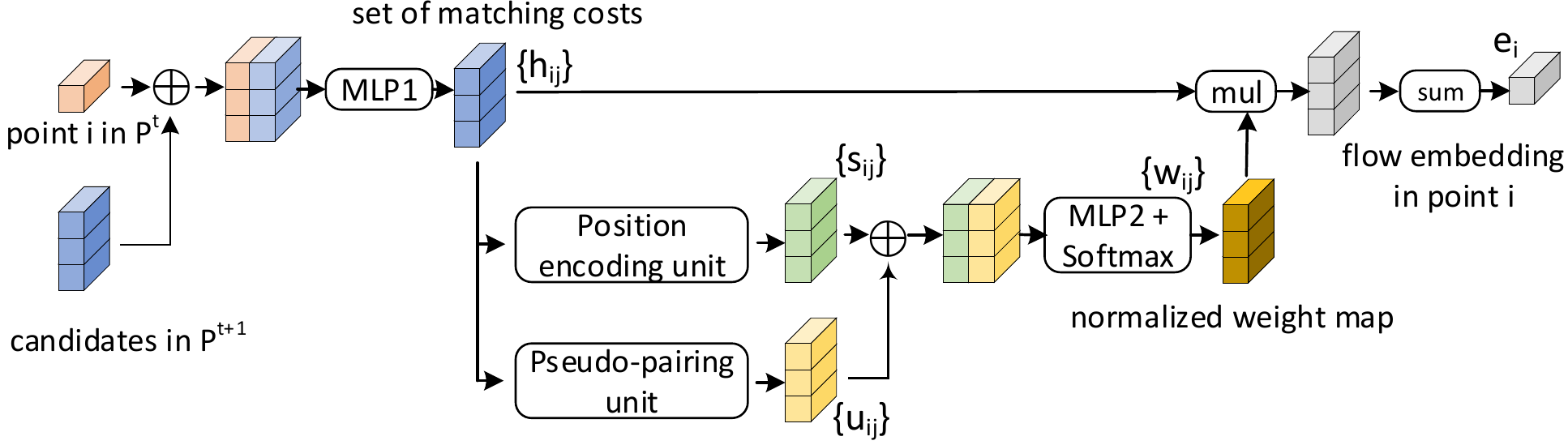}
		\caption{The details of the position-aware flow embedding layer.}
		\label{fig_e}	
	\end{figure}
	
	\section{HCRF-Flow}

	\subsection{Overview}
	
	In the task of point cloud scene flow estimation, the inputs are two point clouds at two consecutive frames: ${\bm P}^t = \{{\bm p}^t_{i} | i = 1,...,n_t\}$ at frame $t$ and ${\bm P}^{t+1} = \{{\bm p}^{t+1}_{j} | j = 1,...,n_{t+1}\}$ at frame $t+1$, where ${\bm p}^t_{i} , {\bm p}^{t+1}_{j} \in \mathbb{R} ^3$ are 3D coordinates of individual points. 
	Our goal is to predict the 3D displacement for each point $i$ in ${\bm P}^t$, which describes the motion of each point $i$ from frame $t$ to frame $t+1$.
	Unless otherwise stated, we use boldfaced uppercase and lowercase letters to denote matrices and column vectors, respectively.
	
	As shown in Fig.~\ref{fig_a}, the HCRF-Flow consists of two components: a PAFE module for  per-point flow estimation and a Con-HCRFs module for refinement.  
	For the PAFE module, we try two different architectures, a single-scale architecture referring to FlowNet3D~\cite{qi2017pointnet} and a pyramid architecture referring to PointPWC-Net~\cite{wu2020pointpwc}.
	To mix the two point clouds, in the PAFE module, we propose a novel position-aware flow embedding layer to build reliable matching costs and aggregate them to produce flow embeddings that encode the motion information.
	For better aggregation, we use the position information and the matching cost difference as clues to generate aggregation weights.	
	Sec.~\ref{sec_em} introduces the details about this layer.
	In the Con-HCRFs module, we propose novel continuous high order CRFs to refine the coarse scene flow by encouraging both point-level and region-level consistency.
	More details are given in Sec.~\ref{sec_chocrf}.

	\subsection{Position-aware flow embedding layer}\label{sec_em}
	
	As shown in Fig.~\ref{fig_e}, the position-aware flow embedding layer aims to produce flow embedding $\bm e_i$ for each point  $\bm p_i$ in $\bm P^t$.
	For each point $\bm p_i$, we first find neighbouring points $\mathcal{N}_e(i)$ around  $\bm p_i^t$ in $t+1$ frame $\bm P^{t+1}$.
	Then, following~\cite{liu2019flownet3d}, the matching cost between point $\bm p_i^t$ and a softly corresponding point $\bm p_j^{t+1}$  in $\mathcal{N}_e(i)$ are addressed as:
	\begin{small}
		\begin{gather}\label{hij}
		\bm h_{ij} = h(\bm  f_i^t,\bm  f_j^{t+1},  {\bm  p}^{t+1}_{j} - {\bm p}^{t}_{i}),
		\end{gather}
	\end{small}
	where $\bm f_i^t$ and $\bm f_{j}^{t+1}$ are the features for $\bm p_i^t$  and $\bm p_j^{t+1}$, respectively. $h$ is a concatenation of its inputs followed by a MLP. 
	After obtaining the matching costs for point $\bm p_i^t$, two subbranches are followed: position encoding unit and pseudo-pairing unit, to produce weights for aggregation, as shown in Fig.~\ref{fig_e}.

	\noindent\textbf{Pseudo-pairing unit}\quad 
	When aggregating the matching costs, this unit is designed to automatically select prominent ones by assigning them more weights.
	To this end, we compare each matching pair with a pseudo stationary pair and use the difference as a clue to measure the importance of this matching pair. 
	The pseudo stationary pair represents the situation when this point does not move, $i.e.$ the softly corresponding point is itself.
	Based on Eq.~\ref{hij}, the matching cost of the pseudo stationary pair for each point $p_i$ can be defined as:
	\begin{small}
		\begin{gather}
		{\bar {\bm  h}}_{i} = h(\bm f_i^t,\bm f_i^{t},  { \bm p}^{t}_{i} - {\bm p}^{t}_{i}).
		\end{gather}
	\end{small}
	The matching cost difference between each matching pair and this pseudo pair can be expressed as:
	\begin{small}
		\begin{gather}
		{{\bm  u}}_{ij} = {\bm  h}_{ij} - {\bar {\bm  h}}_{i}.
		\end{gather}
	\end{small}
	In subsequent aggregation procedure, the matching cost difference ${{\bm  u}}_{ij}$ will be treated as a feature to produce aggregation weights for each matching cost.
	
	\noindent\textbf{Position encoding  unit}\quad 
	Further, to improve the ability of our aggregation, we incorporate the position representations into the aggregation procedure as a significant factor in producing soft weights.
	Specifically, inspired by~\cite{shaw2018self} and~\cite{hu2020randla},  for each matching pair $\bm p_i^t$ and $\bm p_j^{t+1}$, we utilize 3D Euclidean distance, the absolute and relative coordinates as position information to encode the position representation $\bm s_{ij}$, which can be expressed as:
	\begin{small}
		\begin{gather}\label{s_ij}
		\bm s_{ij} = M_s(\bm p_i^t  \oplus \bm p_j^{t+1}   \oplus  (\bm p_i^t -  \bm p_j^{t+1} )  \oplus   \| \bm p_i^t -  \bm p_j^{t+1} \| ),
		\end{gather}
	\end{small}
	where $M_s(\cdot)$  is a MLP to map the position information into the position representation, $\oplus$ is the concatenation operation, and $\| \cdot \|$ computes the Euclidean distance between the two points.
	
	Given the matching cost difference and the position representation, we design a shared function $M_a(\cdot)$ to produce a unique weight vector for each matching cost for aggregation. 
	Specifically, the function $M_a(\cdot)$ is composed of a MLP followed by a \textit{softmax} operation to normalize the weights across all matching costs in a set.
	The normalized weight for each matching cost ${ \bm h}_{ij} $ can be written as:
	\begin{small}
		\begin{gather}
		{ \bm w}_{ij} =M_a(\bm u_{ij}  \oplus \bm s_{ij} ).
		\end{gather}
	\end{small}
	Therefore, for each point ${ \bm p}_{i}$, according to the learned aggregation weights,  the final flow embedding for point $\bm p_i^t$ can be expressed as:
	\begin{small}
		\begin{gather}\
		\bm e_i = \sum\nolimits_{j \in \mathcal{N}_e(i) }{\bm w}_{ij}  \odot {\bm h}_{ij},
		\end{gather}
	\end{small}
	where $\odot $ is the element-wise multiplication and $\mathcal{N}_e(i)$ is the set of softly corresponding points for point $\bm p_i^t$ in next frame.

	\section{Continuous High-Order CRFs}\label{sec_chocrf}
	In this section, we introduce the details of our continuous high order CRFs. 
	We first formulate the problem of scene flow refinement.
	Then, we describe the details of three kinds of potential functions involved in the Con-HCRFs.
	Lastly, we discuss how to utilize mean field theory to approximate the Con-HCRFs distribution and obtain the final iterative inference algorithm.

	\subsection{Overview}
	Take a point cloud $\bm{P}$ with $n$ points, indexed $1,2, ..., n$. 
	In the scene flow refinement, we attempt to assign every point a refined 3D displacement based on the initial scene flow produced by the PAFE module.
	Let $\bm{Y} = [\bm{y}_1, \bm{y}_2, ..., \bm{y}_n]$ be a matrix of 3D displacements corresponding to all $n$ points in point cloud $\bm{P}$, where each $\bm{y}_i \in \mathbb{R} ^3$. 
	Following~\cite{liu2015deep,ristovski2013continuous}, we model the conditional probability distribution with the following density function 
	$
	{\rm Pr}( \bm{Y} | \bm{P} ) = \frac{1}{{\rm Z } (\bm P) } {\rm exp} ( - E( \bm{Y} | \bm{P})). 
	$
	Here $E$ is the energy function and $Z$ is the partition function defined as:
	$
	Z(\bm{P}) = \int_{Y}{{\rm exp} ( - E( \bm{Y} | \bm{P})) }{\rm d}\bm{Y}. 
	$
	
\begin{figure}[tb]
		\centering
		\includegraphics[height=3.2cm]{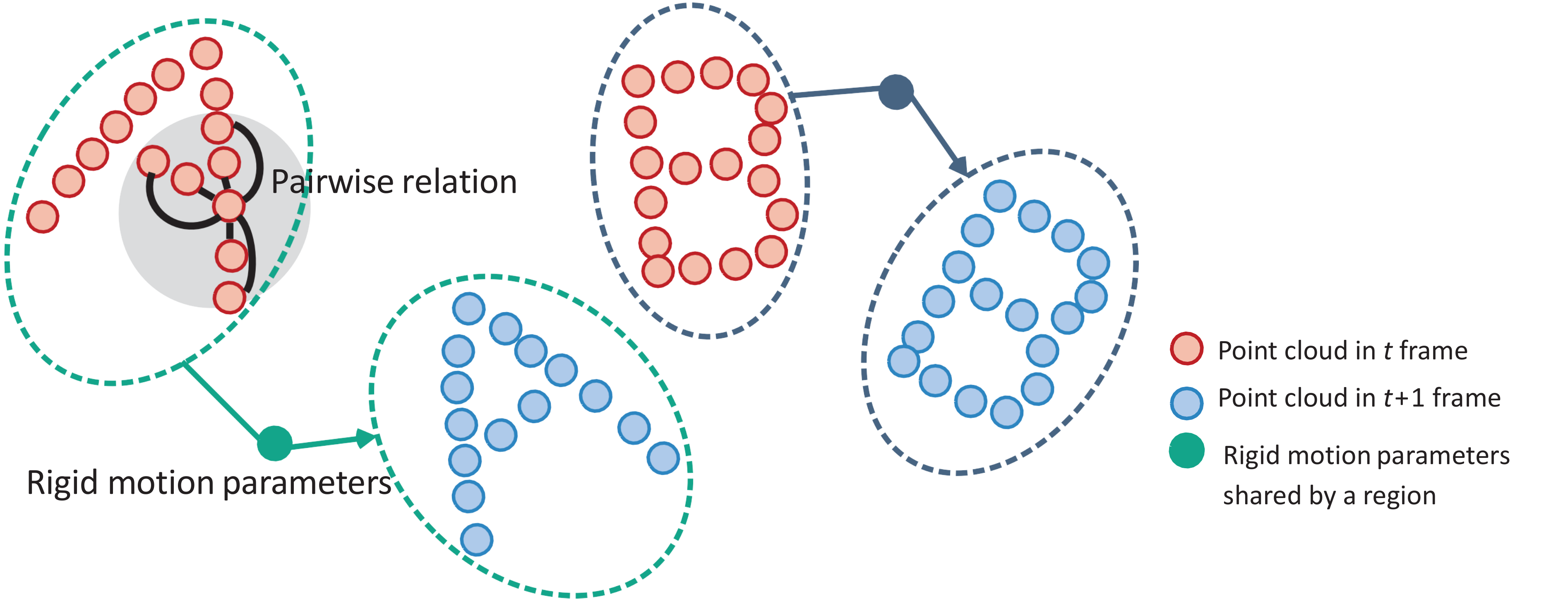}
		\caption{Illustration of Con-HCRFs. Red points and blue points represent the point cloud in frame~$t$ and frame~$t+1$, respectively. 
			The black lines and gray background represent the pairwise relations and the neighborhood in the pairwise term.
			The dashed boxes cover rigid regions. 
			An arrow with a point represents the rigid motion of a region with rigid motion parameters shared by all the points in it. 
			The rigid motion constraints compose the high order term in Con-HCRFs.
		}
		\label{fig_crf}
	\end{figure}

	Different from conventional CRFs, the Con-HCRFs proposed in this paper contains a novel high order potential that performs rigid motion constraints in each local region. 
	Specifically, the energy function is defined as:
	
	\begin{footnotesize}
		\begin{equation}
		\begin{aligned}
		E( \bm{Y} | \bm{P}) = &\sum_{i}^{}{\psi^{U}( \bm{y}_i, \bm{P})  }   
		+   \sum_{i, j \in {\cal{N}}(i) }^{}{\psi^{B}( \bm{y}_i, \bm{y}_j, \bm{P})  }\\
		&+ \sum_{ {\cal{V}} \in {\bm V}_{set} }^{}{  \sum_{ i \in {\cal{V}}}^{}{   \psi^{SV}(\bm{y}_i, \bm{Y}_{{\cal{V}} - i}, \bm{P}) }}, 
		\end{aligned}
		\end{equation}
	\end{footnotesize}
	where ${\cal{N}}(i)$ represents the set of neighboring points of center point $i$; $ {\bm V}_{set} $ represents the set of rigid regions in the whole point cloud and $\bm{Y}_{{\cal{V}} - i}$ is a matrix composed by the scene flow of points belonging to the region ${\cal{V}}$ with the point $i$ excluded.
	And the point set without the point $i$ is denoted as ${{\cal{V}} - i}$.
	The unary term $\psi^{U}$ encourages the refined scene flow to be consistent with the initial scene flow. 
	The pairwise term $\psi^{B}$ encourages neighboring points with similar local structure to take similar displacements. 
	The high order term $\psi^{SV}$ encourages points belonging to the same rigid region to share the same rigid motion parameters.
	In this paper, we use over-segmentation method to segment the entire point cloud into a series of supervoxels and treat each supervoxel as the rigid region in the high order term. 
	An illustration is shown in Fig.~\ref{fig_crf}. 
	We drop the conditioning on $\bm{P}$ in the rest of this paper for convenience.

	\subsection{Potential functions}
	
	\textbf{Unary potential}\quad 
	The unary potential is constructed from the initial scene flow by considering $L^2$ norm:
	\begin{gather}
	\label{unary}
	\psi^{U}( \bm{y}_i) = \| \bm{y}_i - \bm{z}_i\|_2^2,
	\end{gather}
	where $\bm{z}_i$ represents the initial 3D displacement at point $i$ produced by PAFE module; $\| \cdot \|_2$ denotes the $L^2$ norm of a vector.\\
	
	\noindent\textbf{Pairwise potential}\quad 
	The pairwise potential is constructed from $C$ types of similarity observations to describe the relation between pairs of hidden variables $\bm{y}_i$ and $\bm{y}_j$:
	\begin{small}
		\begin{gather}
		\label{pairwise}
		\psi^{B}( \bm{y}_i, \bm{y}_j) = \sum\nolimits_{c =1}^{C} {  \alpha_c K_{ij}^{(c)}  \| \bm{y}_i - \bm{y}_j\|_2^2},
		\end{gather}
	\end{small}
	where $K_{ij}^{(c)}$ is a weight to specify the relation between the points $i$ and $j$; $ \alpha_c$ denotes the coefficient for each similarity measure. 
	Specifically, we set weight $K_{ij}^{(c)}$ depending on a Gaussian kernel $K_{ij}^{(c)} ={\rm \exp}  (- \frac{\| \bm{k}_i^{c} - \bm{k}_j^{c} \|_2^2}{2\theta_c^2})$, where $\bm{k}_i^c$ and $\bm{k}_j^c$ indicate the features of neighboring points $i$ and $j$ associating with similarity measure $c$; $\theta_c$ is the kernel's bandwidth parameter.
	In this paper, we use point position and surface normal as the observations to construct two Gaussian kernels.\\
	
	\noindent\textbf{High order potential}\quad 
	For the high order potential term, we want to explore the effects of interactions among points in a supervoxel.
	According to the rigid motion constraint, the high order potential term in  CRF can be defined as:
	\begin{gather}
	\label{high}
	\psi^{SV}(\bm{y}_i,\bm{Y}_{{\cal{V}} - i})=\beta  \|\bm{y}_i - {\bm{g}}({\bm{p}_i},\bm{Y}_{{\cal{V}} - i})\|_2^2,
	\end{gather}
	where ${\bm{g}}({\bm{p}_i},\bm{Y}_{{\cal{V}} - i})$ is a displacement produced by a function ${\bm{g}}(\cdot,\cdot)$, where shared rigid motion parameters will be computed by $\bm{Y}_{{\cal{V}} - i}$ and the displacement for point $i$ obeying the shared parameters can be obtained by applying the shared parameters back to the original position ${\bm{p}_i}$.
	$\beta$ is a coefficient. 
	In the following, we will give details about the computation of  ${\bm{g}}({\bm{p}_i},\bm{Y}_{{\cal{V}} - i})$.

	In a rigid region $\cal{V}$, given point $i$, we denote the points in region $\cal{V}$ not containing the point $i$ as 
	${\bm P}_{{\cal{V}} - i} = \{ {{\bm p}_j \in {\bm{P}}}  | j \in {{\cal{V}} - i} \}$ and the corresponding 3D displacements as ${\bm Y}_{{\cal{V}} - i} = \{ {{\bm y}_j \in {\bm{Y}}}  | j \in {{\cal{V}} - i} \}$. 
	The warped positions in the next frame can be obtained by adding the scene flow back to the corresponding positions in frame $t$: 
	\begin{gather}
	{\bm D}_{{\cal{V}} - i}= \{{{\bm d}_j(\bm y_j) }= {{\bm p}_j }+ {{\bm y}_j } | j \in {{\cal{V}} - i}  \} .
	\end{gather}
	The possible rigid transformation from $\bm{P}_ {{\cal{V}}\!-\!i}$ to $\bm{D}_{{\cal{V}}\!-\!i}$ can be defined as
	$[\bm{R}, \bm{t}]$ where $\bm{R}\in {\rm SO}(3)$ and $\bm{t} \in \mathbb{R} ^3$. 
	Inspired by the work~\cite{wang2019deep} in point cloud registration, we can minimize the mean-squared error $E^{rigid}(\bm{R}, \bm{t};{\bm Y}_{{\cal{V}} - i})$ to find the most suitable rigid motion parameters $[\bm{R}^*({\bm Y}_{{\cal{V}} - i}), \bm{t}^*({\bm Y}_{{\cal{V}} - i})]$ to describe the motion ${\bm Y}_{{\cal{V}} - i}$:
	
	\begin{footnotesize}
		\begin{gather}
		E^{rigid}(\bm{R}, \bm{t};{\bm Y}_{{\cal{V}}\!-\!i})= \frac{1}{N_{{\cal{V}}\!-\!i}} \sum\nolimits_{j = 1}^{N_{{\cal{V}}\!-\!i}}{\| \bm{R}\bm{p}_j + \bm{t}- {{\bm d}_j(\bm y_j) }  \|_2^2},
		\end{gather}
	\end{footnotesize}
	\begin{footnotesize}
		\begin{gather}
		[\bm{R}^*({\bm Y}_{{\cal{V}}\!-\!i}), \bm{t}^*({\bm Y}_{{\cal{V}}\!-\!i})]= {\rm{arg}} \min_{\bm{R}, \bm{t}}{E^{rigid}(\bm{R}, \bm{t};{\bm Y}_{{\cal{V}}\!-\!i})},
		\end{gather}
	\end{footnotesize}
	where $N_{{\cal{V}}\!-\!i}$ is the number of points in region ${\cal{V}}$ with the point $i$ excluded.
	
	Define centers of $\bm{P}_{{\cal{V}} - i}$ and $\bm{D}_{{\cal{V}}-i}$ as 
	$\overline{\bm p} _{{\cal{V}}-i}= \frac{1}{N_{{\cal{V}}\!-\!i}}  \sum\nolimits_{j = 1}^{N_{{\cal{V}}\!-\!i}}{{\bm p}_j}  $ and 
	$\overline{\bm d}_{{\cal{V}}\!-\!i} = \frac{1}{N_{{\cal{V}}\!-\!i}}  \sum\nolimits_{j = 1}^{N_{{\cal{V}}\!-\!i}}{{\bm d}_j}, $
	respectively.
	Then the cross-covariance matrix $\bm{H}_{{\cal{V}}\!-\!i}$ can be written as:
	$\bm{H}_{{\cal{V}}\!-\!i} = \sum\nolimits_{j =1}^{N_{{\cal{V}}\!-\!i}}{({\bm p}_j - \overline{\bm p}_{{\cal{V}}\!-\!i}) ({\bm d}_j - \overline{\bm d}_{{\cal{V}}\!-\!i})^\top }.$
	Using the the singular value decomposition (SVD) to decompose $\bm{H}_{{\cal{V}}\!-\!i}= \bm{U}_{{\cal{V}}\!-\!i}  \bm{S}_{{\cal{V}}\!-\!i} \bm{V}_{{\cal{V}}\!-\!i}^\top$, we can get the closed-form solutions of  $E^{rigid}_{{\cal{V}}\!-\!i}(\cdot,\cdot)$, written as:
	\begin{small}
		\begin{gather}
		\begin{aligned}
		\bm{R}^*({\bm Y}_{{\cal{V}}\!-\!i}) & = \bm{V}_{{\cal{V}}\!-\!i} \bm{U}_{{\cal{V}}\!-\!i}^\top  \\
		\bm{t}^*({\bm Y}_{{\cal{V}}\!-\!i})  & = - \bm{R}^*({\bm Y}_{{\cal{V}}\!-\!i}) \overline{\bm p}_{{\cal{V}}\!-\!i} + \overline{\bm d}_{{\cal{V}}\!-\!i}.
		\end{aligned}
		\end{gather}
	\end{small}
	When treating the most suitable parameters $[\bm{R}^*({\bm Y}_{{\cal{V}}\!-\!i}), \bm{t}^*({\bm Y}_{{\cal{V}}\!-\!i})]$ as the shared rigid motion parameters by all points in region $\cal{V}$, the displacement that satisfies the rigid motion constraints for point $i$  is given by:
	\begin{gather}
	\label{yiv}
	{\bm{g}}({\bm{p}_i},\bm{Y}_{{\cal{V}}\!-\!i})= \bm{R}^*({\bm Y}_{{\cal{V}}\!-\!i})  {\bm p}_i + \bm{t}^*({\bm Y}_{{\cal{V}}\!-\!i})  - {\bm p}_i.
	\end{gather}
	
	\subsection{Inference}\label{Inference}
	In order to produce the most probable scene flow, we should solve the MAP inference problem for ${\rm Pr}( \bm{Y} ) $.
	Following~\cite{ristovski2013continuous}, we  approximate the original conditional distribution ${\rm Pr}( \bm{Y} ) $ by mean field theory~\cite{koller2009probabilistic}.
	Thus, the distribution ${\rm Pr}( \bm{Y} ) $ is approximated by a product of independent marginals, i.e.,  $Q( \bm{Y} ) = \prod_{i=1}^{n}Q_i( \bm{y}_i )$.
	By minimizing the  KL-divergence between ${\rm Pr}$ and $Q$, the solution for $Q$ can be written as:
	${\rm log} (Q_i ({\bm y}_i ))  = {\rm E}_{j \neq i}[{\rm Pr}( \bm{Y} ) ] + const,$
	where $ {\rm E}_{j \neq i}$ represents an expectation under $Q$ distributions over all variable ${\bm y}_j$ for ${j \neq i}$.

	Following~\cite{ristovski2013continuous}, we represent each ${\rm log} (Q_i ({\bm y}_i))$ as a multivariate normal distribution,  the mean field update for mean ${\bm \dot{\mu}}_i$ and normalization parameter $\sigma_i$ can be written as:
	
	\begin{small}
		\begin{equation}
		\label{sigma}
		\sigma_i= \frac{1}{2(1 + 2 \sum_{c =1}^{C}{  \sum_{j \in {\cal{N}}(i) }{\alpha_c K_{ij}^{(c)}  } } + \beta)},
		\end{equation}
	\end{small}
	\begin{footnotesize}
		\begin{equation}
		\label{mu}
		{\bm \dot{\mu}}_i= 2 \sigma_i ({\bm z}_i \!+\!  
		2 \sum\nolimits_{c =1}^{C}{  \sum\nolimits_{j \in {\cal{N}}(i) }{\alpha_c K_{ij}^{(c)}{\bm \mu}_j   } }  \!+\!      \beta {\bm{g}}({\bm{p}_i},\bm{M}_{{\cal{V}}\!-\!i})),
		\end{equation}
	\end{footnotesize}
	where $\bm{M}_{{\cal{V}}\!-\!i}$ is a set of mean $\bm \mu_j$ for all $j \in {{\cal{V}}\!-\!i}$; $\sigma_i$ is the diagonal element of covariance ${\bm \Sigma}_i$. 
	The detailed derivation of the inference algorithm can be found in  supplementary.
	We observe that there usually exist hundreds of points in a supervoxel, which makes the rigid parameters computed on all points in the supervoxel excluding the point~$i$ vary close to the rigid parameters computed on all points in the supervoxel, i.e., $[\bm{R}^*({\bm M}_{{\cal{V}}\!-\!i}), \bm{t}^*({\bm M}_{{\cal{V}}\!-\!i})]$ is vary close to $[\bm{R}^*({\bm M}_{{\cal{V}}}), \bm{t}^*({\bm M}_{{\cal{V}}})]$.
	Thus, in practice, we approximate ${\bm{g}}({\bm{p}_i},\bm{M}_{{\cal{V}}\!-\!i})$  in Eq.~\ref{mu} with ${\bm{g}}({\bm{p}_i},\bm{M}_{\cal{V}})$, and the approximated mean ${\bm \mu}_i$ is:

		\begin{footnotesize}
		\begin{equation}
		\label{mu2}
		{\bm \mu}_i = 2 \sigma_i ({\bm z}_i +  
		2 \sum\nolimits_{c =1}^{C}{  \sum\nolimits_{j \in {\cal{N}}(i) }{\alpha_c K_{ij}^{(c)}{\bm \mu}_j   } }  +      \beta {\bm{g}}({\bm{p}_i},\bm{M}_{{\cal{V}}})).
		\end{equation}
	\end{footnotesize}
	After this approximation, we only need to calculate  a set of rigid motion parameters for each supervoxel rather than for each point, which greatly reduces the time complexity.
	
	In the MAP inference, since we approximate  ${\rm Pr}$ with $Q$, an estimate of each ${\bm y}_i$ can be obtained by computing the expected value  ${\bm \mu}_i $ of  the Gaussian distribution $Q_i$:
	\begin{small}
		\begin{gather}
		{\bm{y}}_i^* = {\rm{arg}} \max_{\bm{y}_i}{Q_i}(\bm{y}_i ) = {\bm \mu}_i.
		\end{gather}
	\end{small}
	The inference procedure of our Con-HCRFs can be clearly sketched in~Algorithm~{\ref{alg1}}.

	\begin{algorithm}[]
		\caption{Mean field in Con-HCRFs}
		\label{alg1}
		\hspace*{0.02in}{\bf Input:} 
		Coarse scene flow $\bm Z$;Coordinates of point cloud $\bm P$;\\
		\hspace*{0.02in}{\bf Output:} 
		Refined scene flow ${\bm Y}^*$ ;\\
		\hspace*{0.02in}{\bf Procedure:} 
		\begin{algorithmic}[1]
			\STATE {${\bm \mu }_i \gets {\bm z}_i$ for all $i$;}\\
			\COMMENT{ Initialization}
			\WHILE{ not converged} 
			\STATE{Compute $[\bm{R}^*({\bm M}_{\cal{V}}), \bm{t}^*({\bm M}_{\cal{V}})]$ for supervoxel $\cal{V}$;}
			\STATE{${\bm {\hat \mu}}_i \gets   \bm{R}^*({\bm M}_{\cal{V}})  {\bm p}_i + \bm{t}^*({\bm M}_{\cal{V}})  - {\bm p}_i $; }\\
			\COMMENT{ Message passing from supervoxel}
			\STATE{${\bm {\tilde \mu}}_i^{(c)}  \gets \sum\nolimits_{j \in {\cal{N}}(i) }{K_{ij}^{(c)}{\bm \mu}_j } $;}
			\STATE{${{\tilde \sigma}}_i^{(c)}  \gets \sum\nolimits_{j \in {\cal{N}}(i) }{K_{ij}^{(c)}   }$;}
			\\
			\COMMENT{ Message passing from neighboring points}
			\STATE{${\bm {\bar \mu}}_i  \gets {\bm z}_i + 2 \sum_{c}{\alpha_c {\bm {\tilde \mu}}_i^{(c)}  + \beta {\bm {\hat \mu}}_i  }$; }
			\STATE{${{\bar \sigma}}_i  \gets  1 + 2 \sum_{c}{\alpha_c {{\tilde \sigma}}_i^{(c)} }  + \beta$;}		\\
			\COMMENT{ Weighted summing}
			\STATE{\small${\bm \mu }_i  \gets  \frac{1}{{{\bar \sigma}}_i }  {\bm {\bar \mu}}_i   $;} \\
			\COMMENT{Normalizing}
			\ENDWHILE
			\STATE{$\bm{y}_i^* \gets  {\bm \mu }_i $ for all $i$.} 	
		\end{algorithmic}
	\end{algorithm}

	Moreover, thanks to the differentiable SVD function provided by PyTorch~\cite{paszke2017automatic}, the mean field update operation is differentiable in our inference procedure. 
	Therefore, following~\cite{zheng2015conditional, xu2017multi}, our mean field algorithm can be fully integrated with deep learning models, which ensures the end-to-end training of the whole framework.

	\section{Experiments}
	In this section, we first train and evaluate our method on the synthetic FlyingThings3D dataset in Sec.~\ref{exp_1}, and then in Sec.~\ref{exp_2} we test the generalization ability of our method on real-world KITTI dataset without fine-tuning. 
	In Sec.~\ref{exp_3}, we validate the generality of our Con-HCRFs on other networks.
	Finally, we conduct ablation studies to analyze the contribution of each component in Sec.~\ref{exp_4}.
	Note that in the following experiments, there are two different architectures of the PAFE module, the single-scale one denoted as PAFE-S and the pyramid one denoted as PAFE.
	And corresponding HCRF-Flow models are denoted as HCRF-Flow-S and  HCRF-Flow, respectively.\\

	\noindent\textbf{Evaluation metrics.}\quad 
	Let $\bm Y^*$ denote the predicted scene flow, and $\bm Y_{gt}$ be the ground truth scene flow. The evaluate metrics are computed as follows.
	\textbf{EPE3D}(m): the main metric, $\| \bm Y^* - \bm Y_{gt}\|_2$ average over each point.
	\textbf{Acc3DS}(\%): the percentage of points whose EPE3D $<$ 0.05m or relative error $< 5\%$.
	\textbf{Acc3DR}(\%): the percentage of points whose EPE3D $<$ 0.1m or relative error $< 10\%$.
	\textbf{Outliers3D}(\%): the percentage of points whose EPE3D $>$ 0.3m or relative error $> 10\%$.
	\textbf{EPE2D}(px): 2D End Point Error, which is a common metric for optical flow.
	\textbf{Acc2D}(\%): the percentage of points whose EPE2D $<3px$  or relative error $< 5\%$.

	\subsection{Results on FlyingThings3D}\label{exp_1}
	FlyingThings3D~\cite{mayer2016large} is a large-scale synthetic dataset.
	We follow~\cite{gu2019hplflownet} to build  the training set and the test set.
	Our method takes $n = 8, 192$ points in each point cloud as input.
	We train our models on one quarter of the training set (4,910 pairs)  
	and evaluate on the whole test set (3,824 pairs).
	
	Referring to PointPWC-Net~\cite{wu2020pointpwc}, we build a pyramid PAFE module, PAFE, and corresponding HCRF-Flow framework, HCRF-Flow.
	Note that, compared with original architecture in~\cite{wu2020pointpwc}, there are three adjustments in our PAFE: 
	1) we replace the MLPs in level 0 with a set conv~\cite{liu2019flownet3d};
	2) we replace all PointConvs~\cite{wu2019pointconv} with set convs~\cite{liu2019flownet3d};
	3) we replace cost volume layers~\cite{wu2020pointpwc} with our position-aware flow embedding layers. 
	In Con-HCRFs, we utilize the algorithm proposed in~\cite{lin2018toward} for supervoxel segmentation.
	During training, we first train our PAFE with a multi-scale loss function used in~\cite{wu2020pointpwc}. 
	Then we add the Con-HCRFs to PAFE for fine-tuning.
	More implementation details are in supplementary.

	The quantitative evaluation results on the Flyingthings3D are shown in Table~\ref{table_1}.
	We compare our method with four baseline models: FlowNet3D~\cite{liu2019flownet3d}, HPLFlowNet~\cite{gu2019hplflownet},
	PointPWC-Net~\cite{wu2020pointpwc}, and
	FLOT~\cite{puy2020flot}.
		As shown in Table~\ref{table_1}, our PAFE module outperforms the above four methods. 
	Further, adding Con-HCRFs and fine-tuning on FlyingThings3D, the final method, HCRF-Flow, achieves the best performance on all metrics.
	Qualitative results are shown in Fig.~\ref{fig_example}.
	
	\begin{table*}\footnotesize
		
		\caption{ Evaluation results on FlyingThings3D and KITTI Scene Flow 2015. Our model outperforms all baselines on all evaluation metrics. Especially, the good performance on KITTI shows the generalization ability of our method.}
		\label{table_1}
		\renewcommand\arraystretch{1.0}	
		\centering	
		
		\begin{tabular}{l@{\hskip 1.4cm}|l@{\hskip 0.3cm}|c@{\hskip 0.10cm}c@{\hskip 0.10cm}c@{\hskip 0.10cm}c@{\hskip 0.10cm}|c@{\hskip 0.10cm}c}
			\Xhline{1.2pt}
			{Dataset} & {Method} & {EPE3D$\downarrow$} & {Acc3DS$\uparrow$} & {Acc3DR$\uparrow$} & { Outliers3D$\downarrow$}
			& {EPE2D$\downarrow$} & { Acc2D$\uparrow$} \\
			\hline\multirow{6}{0.0cm}{FlyingThings3D} &FlowNet3D~\cite{liu2019flownet3d} & 0.0886 & 41.63 & 81.61 & 58.62 & 4.7142 & 60.10\\
			&HPLFlowNet~\cite{gu2019hplflownet} &  0.0804 & 61.44  & 85.55  & 42.87 & 4.6723& 67.64  \\
			&PointPWC-Net~\cite{wu2020pointpwc} &  0.0588 & 73.79   &  92.76  & 34.24  & 3.2390 & 79.94   \\
			&FLOT~\cite{puy2020flot}  &  0.0520 & 73.20   & 92.70   & 35.70  & - &  -  \\
			&Ours (PAFE module) & 0.0535   &  78.90  & 94.93  & 30.51  & 2.8253 & 83.46    \\
			&Ours (HCRF-Flow) & \bf 0.0488 & \bf 83.37  &  \bf 95.07 &  \bf 26.14 & \bf 2.5652 & \bf  87.04   \\
			\hline\multirow{6}{0.0cm}{KITTI} 
			&FlowNet3D~\cite{liu2019flownet3d} &  0.1069 & 42.77 & 79.78 & 41.38  & 4.3424  & 57.51 \\
			&HPLFlowNet~\cite{gu2019hplflownet} &  0.1169 & 47.83  & 77.76  & 41.03 & 4.8055& 59.38  \\
			&PointPWC-Net~\cite{wu2020pointpwc} &  0.0694 & 72.81   &  88.84  & 26.48  & 3.0062 & 76.73   \\
			&FLOT~\cite{puy2020flot}  &  0.0560 & 75.50   & 90.80   & 24.20  &  - & -   \\
			&Ours (PAFE module) &  0.0646  & 80.29   &  93.47 &  20.24 & 2.4829 & 80.80  \\
			&Ours (HCRF-Flow) &  \bf 0.0531  &  \bf 86.31   &  \bf 94.44  &  \bf 17.97  & \bf 2.0700  & \bf 86.56  \\
			\Xhline{1.2pt}
		\end{tabular}
	\end{table*}
	
	\subsection{Generalization results on KITTI}\label{exp_2}
	KITTI Scene Flow 2015~\cite{menze2018object,menze2015joint}  is a well-known dataset for 3D scene flow estimation.
	In this section, in order to  evaluate the generalization ability of our method, we train our model on FlyingThings3D dataset and test on KITTI Scene Flow 2015 without fine-tuning. 
	And the desired supervoxel size and  the bandwidth parameters in KITTI are the same as those in FlyingThings3D.
	
\begin{figure*}[tb]
		\centering
		\includegraphics[height=4.5cm]{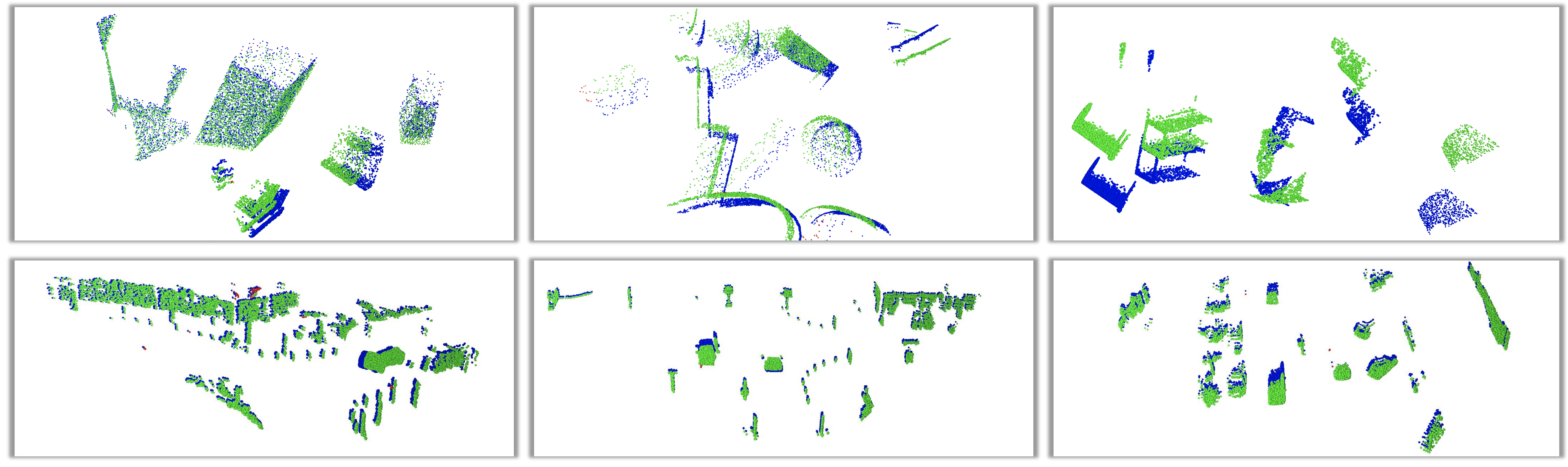}
		\caption{Qualitative results on FlyingThings3D (top) and KITTI (bottom). Blue points are point cloud at frame $t$. Green points are the warped results at frame $t+1$ for the points, whose predicted displacements are  measured as correct by Acc3DR. For the incorrect predictions, we use the ground-truth scene flow to replace them. 
			And  the ground truth warped results are shown as red points.
		}
		\label{fig_example}
	\end{figure*}

	Following~\cite{liu2019flownet3d,gu2019hplflownet}, we evaluate on all 142 scenes in the training set and remove the points on the ground by height ($<0.3m$) for a fair comparison.
	The quantitative evaluation results on the KITTI are shown in Table~\ref{table_1}.
	Our method outperforms the competing methods, which represents the good generalization ability of our method on real-world data.
	Fig.~\ref{fig_example} shows the qualitative results.
	
	\subsection{Generality of Con-HCRFs on other models}\label{exp_3}
	In this section, we study the generalization ability of Con-HCRFs by applying it to the other scene flow estimation models as a post-processing module.
	We evaluate the performance of our proposed Con-HCRFs with FlowNet3D~\cite{liu2019flownet3d} and FLOT~\cite{puy2020flot}, which have shown strong capability on both challenging synthetic data from FlyingThings3D and real Lidar scans from KITTI.
	The results are presented in Table~\ref{table_2}.
	Although built upon strong baselines, our proposed Con-HCRFs boost the performance of each baseline by a large margin on both datasets, demonstrating the strong robustness and generalization.

	\begin{table}\footnotesize
		\caption{ Generalization results of Con-HCRFs on FlowNet3D and FLOT models. $\rm \Delta$ denotes the difference in metrics with respect to each original model.
		Although built upon strong baselines, our proposed Con-HCRFs boost the performance of each baseline by a large margin on the two datasets.}
		\label{table_2}
		\renewcommand\arraystretch{1.0}	
		\centering	
		\begin{tabular}{l@{\hskip 1.5cm}|l@{\hskip 0.15cm}|r@{\hskip 0.15cm}r@{\hskip 0.15cm}}
			\Xhline{1.2pt}
			{Dataset} & {Method}  &  {Acc3DS$\uparrow$} & {$\rm \Delta$Acc3DS}   \\ 
			\hline\multirow{4}{0.0cm}{FlyingThings3D} &FlowNet3D~\cite{liu2019flownet3d} &41.63 &  0.0\\
			&+ Con-HCRFs & 47.01 & + 5.38 \\ \cline{2-4}
			&FLOT~\cite{puy2020flot} & 73.20 & 0.0\\
			&+ Con-HCRFs & 78.63 & + 5.43\\	
			\hline\multirow{4}{0.0cm}{KITTI} &FlowNet3D~\cite{liu2019flownet3d} & 42.77 & 0.0\\
			&+ Con-HCRFs & 46.90 & + 4.13\\ \cline{2-4}
			&FLOT~\cite{puy2020flot} &75.50 & 0.0\\
			&+ Con-HCRFs &  85.44 &  + 9.94\\
			\Xhline{1.2pt}
		\end{tabular}
	\end{table}

	\subsection{Ablation studies}\label{exp_4}
	In this section, we provide a detailed analysis of every component in our method. All experiments are conducted in the FlyThings3D dataset. 
	Besides the pyramid models, PAFE and HCRF-Flow, for comprehensive analysis, we also evaluate the performance of each component when used in single-scale models,  PAFE-S and HCRF-Flow-S. These models are designed referring to FlowNet3D~\cite{liu2019flownet3d}.
	
	\noindent\textbf{Ablation for position-aware flow embedding layer.}\quad 
	We explore the effect of the aggregation strategy on  our position-aware flow embedding layer.
	This strategy is introduced to dynamically aggregate the matching costs  considering position information and matching cost differences.
	As shown in Table~\ref{table_3},  for the two baselines, when applying both the pseudo-pairing unit and the position encoding unit to the flow embedding layer, the performance in Acc3DS can be improved by around 8.
	Moreover, to verify the effectiveness of the pseudo pair, we design a naive dynamic aggregation unit, denoted as~\textbf{NDA},  which directly produces weights from matching costs rather than the matching cost difference between each matching pair and this pseudo pair.
	As shown in Table~\ref{table_3}, after replacing~\textbf{PP} with~\textbf{NDA}, the improvement  in Acc3DS decreases from 6.79 to 2.62.
	Thus, the pseudo-pairing unit is a better choice for this task.

	\begin{table}\footnotesize
		\caption{Ablation study for   position-aware flow embedding layer. 
			\textbf{MP}:~Maxpooling. 
			\textbf{PP}:~pseudo-pairing unit.
			\textbf{PE}:~position encoding unit.
			\textbf{NDA}:~naive dynamic aggregation unit. 
			$\rm \Delta$ denotes the difference in metrics with respect to each baseline model.}
		\label{table_3}
		\renewcommand\arraystretch{1.0}	
		\centering	
		
		\begin{tabular}{l@{\hskip 0.00cm}|c@{\hskip 0.1cm}c@{\hskip 0.1cm}c@{\hskip 0.1cm}c@{\hskip 0.1cm}|r@{\hskip 0.1cm}r@{\hskip 0.0cm}}
			\Xhline{1.2pt}
			{Method}&{MP} & {PP} & {PE} & {NDA} 
			& { Acc3DS$\uparrow$} & { $\rm \Delta$Acc3DS}\\
			\hline
			Single-scale baseline &$\checkmark$ & & &  & 41.63 & 0.00\\
			+NDA  && & &$\checkmark$ & 44.25 & + 2.62\\
			+PP&& $\checkmark$& & & 48.42 & + 6.79\\
			+PP+PE (PAFE-S module) &&$\checkmark$ &$\checkmark$ & & \bf50.08 & + 8.45\\
			\hline
			Pyramid baseline &$\checkmark$ & & &  & 69.94 & 0.0\\
			+PP+PE (PAFE module) &&$\checkmark$ &$\checkmark$ & & \bf 78.90 & + 8.96\\
			\Xhline{1.2pt}
		\end{tabular}
	\end{table}

	\noindent\textbf{Ablation for Con-HCRFs.}\quad 
	To ensure the spatial smoothness and the local rigidity of the final predictions, we propose continuous CRFs with a novel high order term. 
	The ablation results for Con-HCRFs are presented in Table~\ref{table_4}.
	With the help of a pairwise term, denoted as  (Unary+Pair), the  performance gains a slight improvement due to the fact that this pairwise term aims at spatial smoothness but ignores the potential rigid motion constraints. 
	Our proposed Con-HCRFs module, which formulates the rigid motion constraints as its high order term, boosts the performance by a large margin for  PAFE-S and PAFE modules.
	After jointly optimizing Con-HCRFs and PAFE modules, we observe further improvement.
	
	\noindent\textbf{Can we replace the rigid motion constraints by a region-level smoothness in a supervoxel? }\quad 
	We want to explore whether the rigid motion constraint is a good approach to model the relations among points in a supervoxel. 
	Instead of sharing unique rigid motion parameters, a straightforward approach is to encourage the points among a rigid region to share the same motion, i.e., encourage region-level smoothness in a supervoxel.
	To this end, we design a naive regional term as: $	\psi^{naive}(\bm{y}_i, \bm{Y}_{\cal{V}})=\beta  \|\bm{y}_i - {\bm{g}^{naive}}({\bm{p}_i},\bm{Y}_{\cal{V}})\|^2$, where ${\bm{g}^{naive}}({\bm{p}_i},\bm{Y}_{\cal{V}})$ is an average of $\bm{Y}_{\cal{V}}$ over all points in ${\cal{V}}$.
	The results are shown in Table~\ref{table_4}, denoted as (Unary+Pair+naive Region). 
	As it  only enforces spatial smoothness in a region and  fails to model suitable dependencies among points in this rigid region,
	this kind of CRFs is ineffective and even worsen the performance. In contrast, when applying our proposed Con-HCRFs, the final scene flow shows significant improvements.

	\begin{table}\footnotesize
		\caption{Ablation study for  Con-HCRFs. 
			\textbf{Unary}: unary term. \textbf{Pair}: pairwise term. \textbf{High-order}: our proposed high order term.
			\textbf{naive Region}: a naive regional term designed as a reference to verify the effectiveness of our high order term. $\rm \Delta$ denotes the difference in metrics with respect to PAFE or PAFE-S module, whose details are introduced in Table~\ref{table_3}. $\dagger$ means jointly optimizing Con-HCRFs and PAFE modules.}
		\label{table_4}
		\renewcommand\arraystretch{1.0}	
		\centering	
		
		\begin{tabular}{l@{\hskip 0.2cm}|r@{\hskip  0.2cm}r@{\hskip 0.00cm}}
			\Xhline{1.2pt}
			{Method}  & { Acc3DS$\uparrow$} & { $\rm \Delta$Acc3DS} \\ 
			\hline
			PAFE-S module  & 50.08    & 0.00\\
			+ (Unary+Pair) &  50.24 & + 0.16\\
			+ (Unary+Pair+naive Region)  & 48.10   & - 1.98\\
			+ (Unary+Pair+High-order)/Con-HCRFs   &  54.51  & + 4.43\\
			+ (Unary+Pair+High-order)/Con-HCRFs$^\dagger$ &  \bf 56.29 &   + 6.21 \\
			\hline
			PAFE module &  78.90  & 0.00\\
			+ (Unary+Pair+High-order)/Con-HCRFs  & 81.39  & + 2.49\\
			+ (Unary+Pair+High-order)/Con-HCRFs$^\dagger$ & \bf 83.37&  + 4.47 \\
			\Xhline{1.2pt}
		\end{tabular}
	\end{table}

	\begin{table}\footnotesize

		\caption{ Time consumption of Con-HCRFs.}
		\label{table_time}
		\renewcommand\arraystretch{1.0}	
		\centering	
		
		\begin{tabular}{c@{\hskip 0.1cm}|c@{\hskip 0.1cm}|c@{\hskip 0.1cm}|c@{\hskip 0.1cm}|c}
			\Xhline{1.2pt}
			{Component} & {Supervoxel}  & {Pairwise term} & {High order term} & {Total} \\
			\hline{Time (ms)} & {115.1}  & {12.3} & {100.8} & {228.2} \\
			\Xhline{1.2pt}
		\end{tabular}
	\end{table}

	\noindent\textbf{Speed analysis of Con-HCRFs}\quad 
	Table~\ref{table_time} reports the average runtime of each component of Con-HCRFs tested on a single 1080ti GPU.  
	As shown in Table~\ref{table_time}, the Con-HCRFs takes 0.2s to process a scene with 8192 points. The speed of Con-HCRFs is  similar to DenseCRF~\cite{krahenbuhl2011efficient}, which also takes about 0.2s to process a 320x213 image. 
	Additionally, due to the approximate computation that we apply in the high order term, this term only takes 0.1s for a scene.  
	In contrast, the time for the term will dramatically increase from 0.1s to 14s, if the rigid motion parameters are calculated for each point instead of supervoxel. 
	The large gap of runtime shows that the approximation discussed in Sec.~\ref{Inference} can significantly boost the efficiency of our Con-HCRFs. 
	
	\begin{table}\footnotesize
		\caption{ The impact of point number of each supervoxel on our method.}
		\label{table_6}
		\renewcommand\arraystretch{1.0}	
		\centering	
		\begin{tabular}{c@{\hskip 0.1cm}|c@{\hskip 0.2cm}c@{\hskip 0.2cm}c@{\hskip 0.2cm}c@{\hskip 0.1cm}|c}
			\Xhline{1.2pt}
			{Desired point number} & {80}  & {100} & {140} & {200}&{PAFE-S}\\
			\hline
			{EPE3D$\downarrow$} & {0.0804}  & {0.0788} & {\bf 0.0782} & {0.0790}
			& {0.0815}\\ 
			\Xhline{1.2pt}
		\end{tabular}
	\end{table}
	
	\noindent\textbf{Impact of supervoxel sizes.}\quad 
	To illustrate the sensitivity to supervoxel sizes, we  test our method when facing supervoxels with different point numbers.
	As shown in Table~\ref{table_6}, the method achieves the best performance when the desired point number of each supervoxel is set to a range of 140 to 200.

	\section{Conclusions}
	In this paper, we have proposed a novel point cloud scene flow estimation method, termed HCRF-Flow, by incorporating  the strengths of DNNs and CRFs to perform translational motion regression on each point and operate refinement with both pairwise and region-level regularization.
	Formulating the rigid motion constraints as a high order term, we propose a novel high-order CRF based relation module (Con-HCRFs) considering both point-level and region-level consistency.
	In addition, we design a position-aware flow estimation layer for better matching cost aggregation.  
	Experimental results on FlyingThings3D and KITTI datasets show that our proposed method performs favorably against comparison methods.
	We have also shown the generality of our Con-HCRFs on other point cloud scene flow estimation methods.
	
	\section{Acknowledgements} 
This research was conducted in collaboration with SenseTime. 
This work is supported by A*STAR through the Industry Alignment Fund - Industry Collaboration Projects Grant.
This work is also supported by the National Research Foundation, Singapore under its AI Singapore Programme (AISG Award No: AISG-RP-2018-003), and the MOE Tier-1 research grants: RG28/18~(S) and RG22/19~(S).
	
	{\small
		\bibliographystyle{ieee_fullname}
		\bibliography{egbib}
	}
	
\end{document}